\documentclass[letterpaper, 10 pt, conference]{ieeeconf}  % Comment this line out if you need a4paper

\IEEEoverridecommandlockouts                              % This command is only needed if 
\usepackage{cite}
\usepackage{amsmath,amssymb,amsfonts}
\usepackage{algorithmic}
\usepackage{graphicx}
\usepackage{textcomp}
\usepackage{xcolor}
\usepackage{float}
\usepackage{subcaption}
\usepackage{hyperref}
\usepackage{makecell}
\usepackage{comment}
\title{\LARGE \bf
CARIS: A Context-Adaptable Robot Interface System for Personalized and Scalable Human-Robot Interaction
}
\author{Felipe Arias Russi$^{1}$, Yuanchen Bai$^{2}$ and Angelique Taylor$^{2}$% <-this % stops a space
\thanks{*This material was supported by the National Science Foundation under Grant No. IIS-2423127.}% <-this % stops a space
\thanks{$^{1}$Felipe Arias Russi is with the Department of Systems and Computing Engineering, Mathematics, Universidad de los Andes, Cra 1 Nº 18A - 12, Bogotá D.C., Colombia
        {\tt\small af.ariasr@uniandes.edu.co}}%
\thanks{$^{2}$Yuanchen Bai and Angelique Taylor are with the Department of Information Science, Cornell Tech, 2 W Loop Rd, New York, NY 10044, USA
        {\tt\small yb299@cornell.edu, amt298@cornell.edu}}%
}

\begin{document}

\maketitle
\thispagestyle{empty}
\pagestyle{empty}

\begin{abstract}
The human-robot interaction (HRI) field has traditionally used Wizard-of-Oz (WoZ) controlled robots to explore navigation, conversational dynamics, human-in-the-loop interactions, and more to explore appropriate robot behaviors in everyday settings.
However, existing WoZ tools are often limited to one context, making them less adaptable across different settings, users, and robotic platforms. 
To mitigate these issues, we introduce a Context-Adaptable Robot Interface System (CARIS) that combines advanced robotic capabilities such teleoperation, human perception, human-robot dialogue, and multimodal data recording. 
Through pilot studies, we demonstrate the potential of CARIS to WoZ control a robot in two contexts: 1) mental health companion and as a 2) tour guide.
Furthermore, we identified areas of improvement for CARIS, including smoother integration between movement and communication, clearer functionality separation, recommended prompts, and one-click communication options to enhance the usability wizard control of CARIS. 
This project offers a publicly available, context-adaptable tool for the HRI community, enabling researchers to streamline data-driven approaches to intelligent robot behavior.

%various personalized and scalable data collection of HRIs that can increase data-driven approaches in the field.

%The human-robot interaction (HRI) field has traditionally tested robot navigation, conversational dynamics, human-in-the-loop interactions, and more using used Wizard-of-Oz control. However, existing tools are often limited to one context, making them less adaptable across different settings, users, and robotic platforms.  To mitigate these issues, we introduce a Context-Adaptable Robot Interface System (CARIS) for personalized and scalable HRI that combines advanced capabilities like navigation, real-time context understanding and annotation, people tracking, and multimodal data recording. Through pilot studies, we identified areas for improvement for CARIS, including smoother integration between movement and communication, clearer functionality separation, recommended prompts, and one-click communication options to enhance the usability of Large Language Model features. This project offers a publicly available, context-adaptable tool for the HRI community, enabling various personalized and scalable HRIs that can enable robots to learn from prior interactions.
\end{abstract}

% \begin{IEEEkeywords}
% Context-adaptable interface, Large language model, user-friendly design, real-time context, Human-robot interaction.
% \end{IEEEkeywords}

\section{Introduction}

The human-robot interaction (HRI) field has long explored intelligent systems that complement human skills in both social and task-oriented contexts \cite{biermann_how_2021}. 
Social robots must seamlessly coordinate perception, conversation, navigation, and other high-level functions to engage naturally with humans.
This is a demanding task that requires specialized frameworks, diverse toolsets, and depends on robust communication protocols to enable efficient, real-time exchanges between the robot control system and people.
However, prototyping new HRIs remains costly in terms of time, money, and human resources.% (for programming and testing).
Thus, deepening research insights, saving development time, and tailoring systems to specific contexts are essential \cite{dow_wizard_2005}. %To integrate robots into human spaces, prototyping robots cost-effectively is essential for deepening research insights, saving development time, and tailoring systems to specific contexts \cite{dow_wizard_2005}. 
Wizard-of-Oz (WoZ) interfaces have become an effective strategy for iterative system refinement. By employing a human operator to simulate autonomous robot behavior, WoZ control enables rapid prototyping and testing of interaction strategies, offering a flexible and efficient approach to enhance system performance before committing extensive resources to build new interactions \cite{dow_wizard_2005, riek_wizard_2012}.

\begin{figure}[t]
  \centering
  \includegraphics[width=1.0\linewidth]{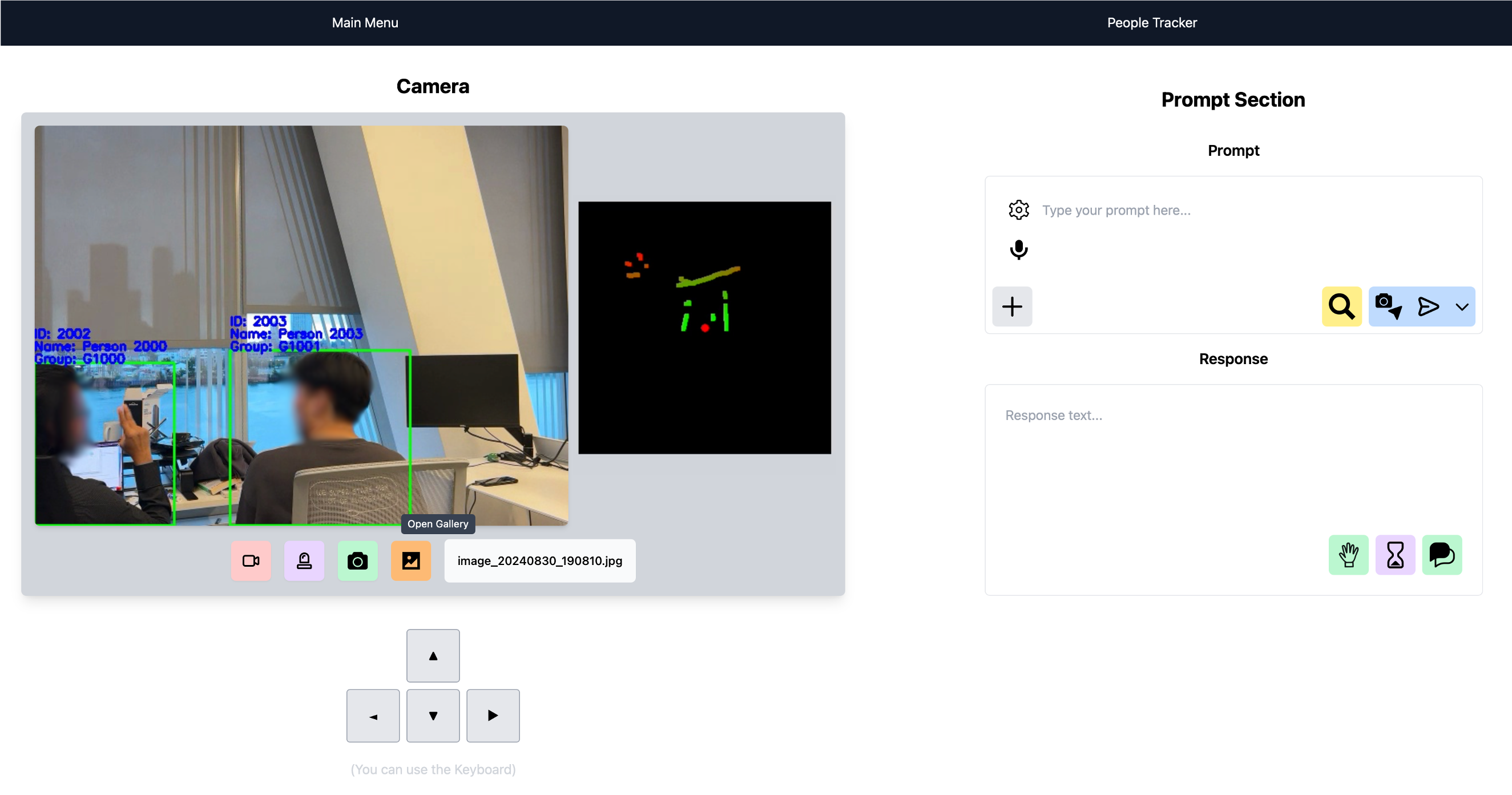}
  \caption{The CARIS Web Interface is a Wizard-of-Oz platform with perception, navigation, conversation modules.}%CARIS Web Interface. \textbf{Top:} Main menu displaying video streams, LIDAR map, teleoperation, and prompt-based LLM interactions, gallery of photos for prompts. \textbf{Bottom:} People Tracking Menu, enabling real-time identification, tagging, and management of detected people, and LLMs.}
  \label{fig:caris-web-interface}
  \vspace{-2pt}
\end{figure}

Although WoZ interfaces can greatly aid prototyping and testing for social robots, developing them from the ground up often requires significant technical effort—something not all HRI teams may find essential to their goals. 
In some cases, existing solutions are tied to specific platforms or robot tasks, and contexts \cite{rietz_woz4u_2021}. 
When HRI researchers explore similar interactions, but under different conditions or with different robotic platforms, they may need to modify these systems substantially. A more flexible approach could help streamline the process by focusing on the core interaction features rather than rebuilding an entire WoZ setup.

Advancements in hardware and container-based workflows \cite{lumpp_design_2024} have led to an array of WoZ interfaces that facilitate seamless integration and interoperability between system components and robots such as perception \cite{ahn_as_2022, karli_alchemist_2024}, navigation \cite{li_tod4ir_2022}, dialogue \cite{barmann_incremental_2024}, computer vision \cite{barmann_incremental_2024, kim_understanding_2024, williams_scarecrows_2024}. 
By leveraging these tools, researchers can rapidly iterate on HRI experiments, deepen their understanding of the operational environment, and reduce the engineering overhead typically associated with building WoZ interfaces from scratch. Nonetheless, while these advances have improved component integration, a unified, general-purpose interface for robots operating in diverse contexts remains lacking.

To address these gaps, we present the Context-Adaptable Robot Interface System (CARIS), an open-source platform that emphasizes ease of modification and streamlined integration with a series of modules. 
CARIS decouples essential functionalities—localization,  perception, and communication into separate modules that can be easily adapted or replaced, offering a flexible tool that enables rapid prototyping for general robots across contexts. 
A unique benefit of CARIS is that these modules kept distinct and containerized so that users can quickly adapt CARIS to different robots and contexts, reusing the same core interaction features with minimal reconfiguration, and real time multimodal data logging to enable HRI researchers to leverage data-driven, machine learning algorithms for HRI.

Our \textbf{research contributions are threefold}. 
First, we introduce CARIS, a context-adaptable WoZ system that enables rapid prototyping and testing of HRI studies. 
Second, CARIS separates essential robot functions from higher-level computing to low-level perception, navigation, conversation, and data recording. 
Third, we conducted experiments to investigate the effectiveness of CARIS after iterative improvements which demonstrate its modularity and adaptability across tour guide and mental health check scenarios.%, we show that CARIS supports rapid HRI evaluations.  and can be refined without rebuilding an entire control interface for new HRIs.

\section{Related Work}

In the HRI field, WoZ prototyping has played a pivotal role to enable rapid iterations of interaction strategies without fully autonomous behavior. Early foundational efforts by Dow et al. \cite{dow_wizard_2005} and Riek \cite{riek_wizard_2012} demonstrated that WoZ experiments can refine both technical and social aspects of robot design, showing that a human operator simulating robot intelligence can reveal insights about user expectations and system requirements. More recent work emphasizes flexible, general-purpose WoZ platforms that can adapt across different robots and interaction settings, rather than being limited to a single hardware or software stack. Table \ref{tab:comparison} summarizes some key characteristics of WoZ systems, selected because they present functionalities or goals related to CARIS.

\begin{table}[!ht]
\centering
%\small
\tiny
\setlength{\tabcolsep}{1pt} % reduce horizontal padding
\begin{tabular}{l c c c c c c c} % Changed first column to left-align
\hline
\textbf{System} & \makecell{\textbf{Speech/Text}\\\textbf{Inputs}} & \makecell{\textbf{LLM}\\\textbf{Integration}} & \makecell{\textbf{Teleoperation}} & \makecell{\textbf{Data}\\\textbf{Storage}} & \makecell{\textbf{Video}\\\textbf{Streaming}} & \makecell{\textbf{Perception}\\\textbf{(SLAM)}} & \makecell{\textbf{Data}\\\textbf{Annotation}} \\
\hline
WoZ4U \cite{rietz_woz4u_2021}              & \textcolor{black}{$\checkmark$} & \textcolor{black}{$\times$} & \textcolor{black}{$\checkmark$} & \textcolor{black}{$\checkmark$}    & \textcolor{black}{$\checkmark$} & \textcolor{black}{$\times$} & \textcolor{black}{$\times$} \\ % Data Storage updated to Check (config file, data recording)
SCOUT \cite{lukin_scout_2024} & \textcolor{black}{$\checkmark$} & \textcolor{black}{$\times$} & \textcolor{black}{$\checkmark$} & \textcolor{black}{$\times$} & \textcolor{black}{$\sim$}     & \textcolor{black}{$\checkmark$} & \textcolor{black}{$\checkmark$} \\ % Data Storage updated to X (runtime storage not explicit)
OpenWoZ \cite{hoffman_openwoz_2016}          & \textcolor{black}{$\checkmark$} & \textcolor{black}{$\times$} & \textcolor{black}{$\checkmark$} & \textcolor{black}{$\times$}    & \textcolor{black}{$\times$}    & \textcolor{black}{$\times$} & \textcolor{black}{$\times$} \\ % Verified based on info
Scarecrows in Oz \cite{williams_scarecrows_2024} & \textcolor{black}{$\checkmark$} & \textcolor{black}{$\checkmark$} & \textcolor{black}{$\times$} & \textcolor{black}{$\times$}    & \textcolor{black}{$\times$}    & \textcolor{black}{$\times$} & \textcolor{black}{$\times$} \\ % Verified based on info
RISE \cite{gros_rise_2023}                & \textcolor{black}{$\checkmark$} & \textcolor{black}{$\checkmark$} & \textcolor{black}{$\checkmark$} & \textcolor{black}{$\checkmark$} & \textcolor{black}{$\times$}    & \textcolor{black}{$\times$} & \textcolor{black}{$\times$} \\ % LLM updated to Check (ChatGPT4all mentioned), Perception to X (SLAM not explicit)
FLEX-SDK \cite{alves-oliveira_flex-sdk_2022}         & \textcolor{black}{$\checkmark$} & \textcolor{black}{$\times$} & \textcolor{black}{$\checkmark$} & \textcolor{black}{$\checkmark$}    & \textcolor{black}{$\times$}    & \textcolor{black}{$\times$} & \textcolor{black}{$\times$} \\ % Verified based on info (Realtime DB counts for Storage)
Malvido-Fresnillo et al.\cite{malvido_fresnillo_open_2024} & \textcolor{black}{$\checkmark$} & \textcolor{black}{$\times$} & \textcolor{black}{$\checkmark$} & \textcolor{black}{$\checkmark$}    & \textcolor{black}{$\checkmark$} & \textcolor{black}{$\times$} & \textcolor{black}{$\times$} \\ % Perception updated to X (SLAM not explicit)
Glauser et al. \cite{glauser_how_2023} & \textcolor{black}{$\checkmark$} & \textcolor{black}{$\sim$} & \textcolor{black}{$\checkmark$} & \textcolor{black}{$\times$}    & \textcolor{black}{$\times$}    & \textcolor{black}{$\checkmark$} & \textcolor{black}{$\times$} \\ % LLM updated to ~ (Dialogflow), Data Storage X, Video X, Perception Check (localization mentioned)
\textbf{CARIS (Ours)}                   & $\checkmark$ & $\checkmark$ & $\checkmark$ & $\checkmark$ & $\checkmark$ & $\checkmark$ & $\checkmark$ \\
\hline
\end{tabular}
\caption{Comparison of CARIS with prior systems. Features assessed include interaction inputs, LLM use, robot control, data handling, and perception/localization. $(\checkmark)$ indicates support, $(\times)$ indicates lack of support, and $(\sim)$ indicates partial or alternative support.}% (e.g., Dialogflow instead of generative LLM; images on demand instead of live video).}
\label{tab:comparison}
\end{table}

%, even if they address different problems or have different objectives from ours.}

Notably, WoZ4U \cite{rietz_woz4u_2021}, RISE \cite{gros_rise_2023}, and the work by Glauser et al. \cite{glauser_how_2023} share a similar goal with CARIS to facilitate HRI experiments, particularly for users who may not have deep technical expertise. Each addresses this challenge through its own architectural choices and feature set. Our work takes a unique approach within this space. %\textcolor{black}{Some systems specifically focus on facilitating WoZ studies. 
WoZ4U \cite{rietz_woz4u_2021}, for example, offers a configurable web interface designed to simplify the execution of WoZ experiments, demonstrated on the Pepper robot. OpenWoZ \cite{hoffman_openwoz_2016} proposes a runtime-configurable WoZ framework architecture, designed for flexibility in HRI research environments. These tools provide valuable means for controlling robots in controlled studies.

Other frameworks aim to support HRI development more broadly. RISE \cite{gros_rise_2023} provides architecture using Robot Operating System (ROS) oriented towards reproducible management of HRI scenarios, supporting module integration for WoZ studies. FLEX-SDK \cite{alves-oliveira_flex-sdk_2022}, is primarily a development kit focused on creating social robot faces and generating speech using text-to-speech software on mobile devices.
%Each offers solutions adapted to their respective research but offers no support for speech generation, teleoperation, video streaming, and data logging.

%while offering tools for interface design and WoZ control, is primarily a development kit focused on creating social robot faces and generating speech. Each offers solutions adapted to their respective research and development niches.

Interfaces designed to improve the general usability of robotic systems also exist. Malvido-Fresnillo et al. \cite{malvido_fresnillo_open_2024} present a reconfigurable web GUI that manages complex ROS-based systems and simplify robot control and monitoring. %, thereby addressing ROS's learning curve. 
Glauser et al. \cite{glauser_how_2023} offer RobotWebControl as part of a middleware that abstracts robot commands via a web interface to facilitate robot programming in healthcare settings.%, specifically facilitating programming for social robotics in healthcare settings.

Finally, some works concentrate on specific aspects of interaction or architecture. SCOUT \cite{lukin_scout_2024} focuses on natural language interaction for collaborative navigation tasks, using WoZ data to train its dialogue models. On the other hand, Scarecrows \cite{williams_scarecrows_2024} conceptually explores LLM integration as modular components within robot architectures to accelerate research.%, drawing an analogy with the utility of WoZ.}

Prior work offers various solutions to streamline WoZ studies for HRI, but CARIS uniquely combines all these efforts through speech generation using Large Language Models, teleoperation, video streaming, data logging, perception, and navigation. %\textcolor{black}{Against this diverse landscape, 
CARIS is presented as a WoZ system with a focus on \textit{context adaptability} through a \textit{modular and integrated} architecture. Its objective is to offer a platform that combines a specific set of capabilities—including SLAM-based localization, real-time perception (such as person tracking), flexible conversation via LLMs, teleoperation, and data handling (see Table \ref{tab:comparison})—accessible from a unified web interface. By integrating these functionalities, CARIS aims to facilitate rapid prototyping and HRI evaluation for researchers across varied scenarios, enabling adaptation to different robots and application contexts without building a new interface. %or building one from scratch.

\section{Context-Adaptable Robot Interface System}
\label{sec:caris}

The core architecture of CARIS is built around three main components: (1) a \emph{robot onboarding} platform that integrates an RGB camera, a 2D LIDAR, and audio (for text-to-speech TTS) on a mobile base capable of moving in two dimensions; (2) a backend implemented in FastAPI, responsible for computer vision tasks and LLM-based requests; and (3) a web interface serving as the WoZ module, built with Svelte and operated by a human wizard. While this three-part division clarifies the system’s technical structure, CARIS is best understood functionally by focusing on four transversal modules—\emph{Localization}, \emph{Perception}, \emph{Conversation}, and \emph{Wizard}—each contributing to the adaptability and rapid prototyping of HRIs in various contexts. 

\begin{figure}[t]
  \centering
  \includegraphics[width=1.0\linewidth]{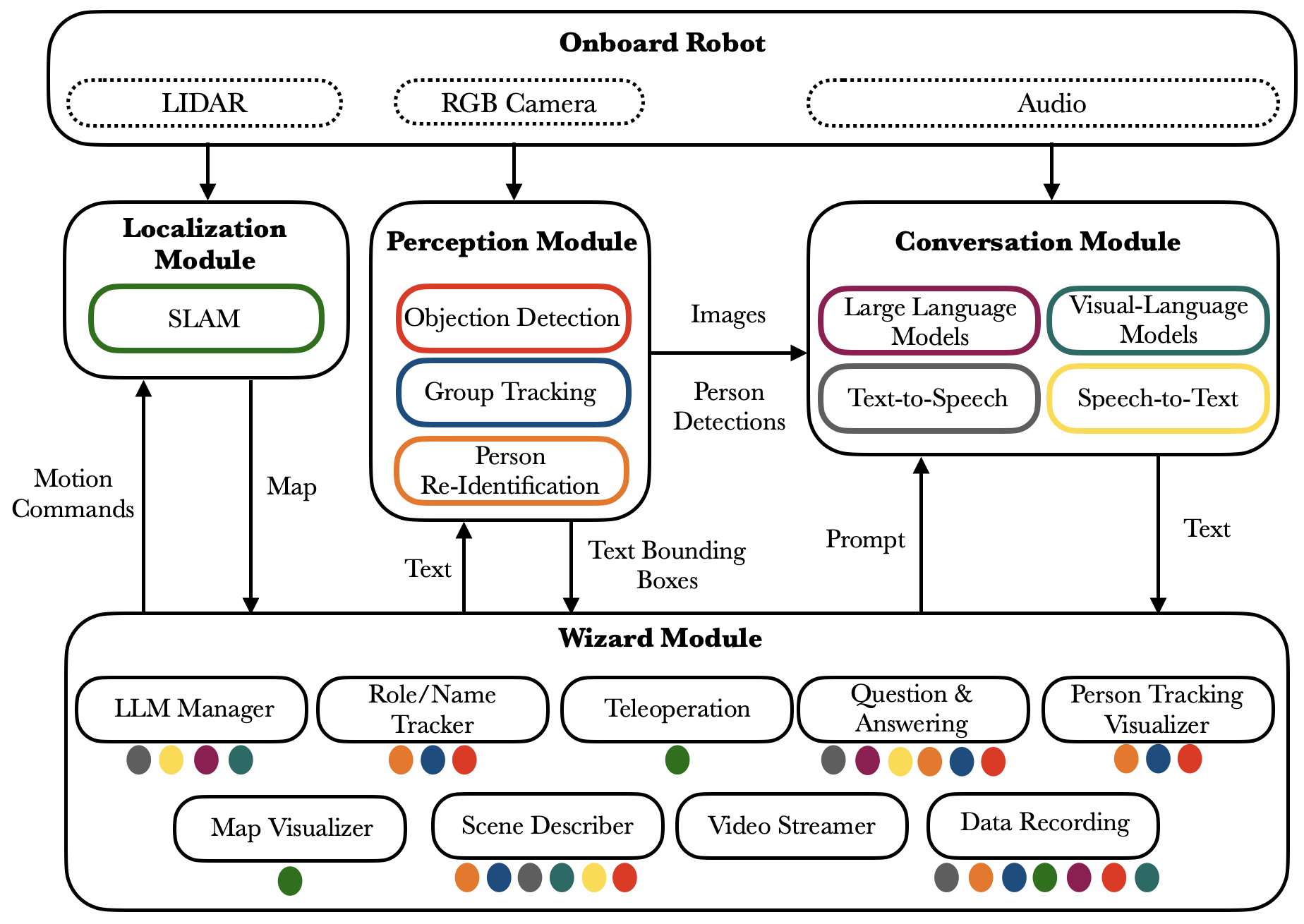}
  \caption{CARIS architecture with LIDAR, RGB camera, audio inputs; Localization, Perception, Conversation, and Wizard Modules.}
  \label{fig:caris-architecture}
  \vspace{-2pt}
\end{figure}

The modularity of CARIS allows researchers to decouple hardware components and repurpose the system for different robots without extensive reconfiguration. Figure~\ref{fig:caris-architecture} provides a structural overview of CARIS, illustrating the independent yet interconnected nature of the robot's onboard components, backend processing, and Wizard interface. This design enables seamless adaptation from lab settings or public spaces.%to diverse HRI studies, whether in lab settings or public spaces.

\subsection{Localization Module}
The \emph{Localization Module} manages the robot’s movement and spatial awareness using a 2D LIDAR and Simultaneous Localization and Mapping (SLAM) techniques. Through ROSBridge, the web interface (the \emph{Wizard Module}) can issue movement commands (forward, backward, rotate left, and rotate right) without needing direct access to the robot’s control stack. This loose coupling makes it easier to change sensors and teleoperation for different mobile platforms. %This loose coupling makes it easier to switch between different sensors and navigation strategies for various types of mobile platforms.

%Besides localization and obstacle avoidance, the \emph{Navigation Module} plays a key role in adapting a robot to different contexts. For example, in one-on-one interactions, the robot might approach a user and maintain a comfortable distance by dynamically recalculating paths. In group scenarios, it can navigate among multiple people without risking collisions, while in remote teleoperation, the module streams a live map and sensor data to a distant operator. Thus, minimal adjustments are needed to switch between in-person social engagements and remote telepresence applications.

\subsection{Perception Module}

The \emph{Perception Module} oversees object detection, tracking, and re-identification. In the current implementation, it leverages \emph{YOLO v8n} \cite{jocher_yolov8_2023} and \emph{DeepSORT} \cite{wojke_simple_2017} to detect and follow individuals in real time via the robot’s RGB camera. However, these algorithms can be reconfigured to handle other types of objects if needed.
This functionality in the backend (under FastAPI) brings significant advantages. First, resource-intensive operations—such as neural network inference—are offloaded from the core interaction loop, preventing delays. Second, the real-time tracking and labeling of individuals can be immediately reflected in the \emph{Wizard Module}, enabling the operator to rename or group detected people (see Fig~\ref{fig:caris-web-interface}). The system can also “remember” labeled users and retrieve their interaction histories later. This is crucial for social scenarios where the robot may need to engage with the same individuals long-term. %over multiple encounters or adapt to new environments without losing previously gathered context (e.g., from a lab setting to a lobby area).

\subsection{Conversation Module}
The \emph{Conversation Module} handles all language-based interactions. On the input side, it provides Speech-to-Text (STT) to capture spoken requests or commands from the wizard to the end-user. %operator or a remote user. On the output side, 
Text-to-Speech (TTS) broadcasts responses through the robot’s speakers. In between, it connects to different LLMs, capable of handling text input alone or combined with images (\emph{Visual Language Models}) for richer interactions. By default, the backend supports cloud-based language models such as Gemini Flash 1.5 \cite{google_gemini_2025} and local models like Llama 3.1 8B \cite{meta_llama_2024} or LLaVA 7B \cite{liu_visual_2023}. CARIS offers flexibility in deployment environments, whether for low-latency local inference or more powerful remote servers.%This variety offers flexibility in deployment environments, whether one needs low-latency local inference or more powerful remote servers.

Similar to the Perception Module, the Conversation Module resides in the backend and communicates with the web interface via HTTP requests. This design allows users to store, retrieve, and repurpose captured images (e.g., for object recognition or to contextualize queries to an LLM). It also supports dynamic role assignments or “interaction styles” for the language model, making the dialog system more adaptable. In \emph{telepresence} use cases, a remote operator could speak through the robot using this module, effectively giving the distant user a “voice” in the local environment. In both individual and group interactions, the system can keep track of who is speaking and in what context—merging data from the \emph{Perception Module} and the \emph{Conversation Module} to sustain coherent, context-aware dialogues.

%By default, the backend supports cloud-based language models such as \emph{Gemini Flash 1.5} and local models like \emph{Llama 3.1} and \emph{Ollava}. This variety offers flexibility in deployment environments, whether one needs low-latency local inference or more powerful remote servers.

\subsection{Wizard Module}
The \emph{Wizard Module} is the web-based interface, developed in Svelte, and serves as the central orchestrator, coordinating information exchange with the other modules. Its main role is to provide the user or researcher with a control panel (see Fig.~\ref{fig:caris-web-interface}) that brings together:
\begin{itemize} 
    \item \textbf{Teleoperation}: Sends movement commands to the \emph{Localization Module}. 
    \item \textbf{Real-time visualization}: Displaying live camera streams, LIDAR maps, person and object detection. %and identifications of people or objects. 
    \item \textbf{Interaction and data management}: Tags, renames, and groups detected people; captures and reviews conversation histories; logs movement commands and TTS usage as text files; stores photos as PNG files; and records LLM interactions as JSON files.
    
    \item \textbf{Dialogue control}: Sends prompts to LLMs (text and images), sets roles or default prompts, enables or disables STT/TTS, and uses conversation templates. 
\end{itemize}

By unifying these functionalities, the \emph{Wizard Module} acts as a “command center,” allowing users to supervise and drive the robot’s capabilities without advanced robotics or programming knowledge. In iterative WoZ-based experiments, this modular framework supports rapid testing of new interaction strategies, local data storage, and straightforward integration of updated algorithms—whether for perception, or conversation—without overhauling the entire interface.

Overall, breaking CARIS down into four interconnected modules not only clarifies the architecture but also underscores the aim of a \emph{multifunctional, easily adaptable} system for diverse social robotics scenarios. Each module takes on specific responsibilities and seamlessly integrates with the others, enabling a robot to transition from focused one-on-one to group interactions with minimal configuration.%, say, a focused one-on-one interaction to a broader public demonstration with minimal additional configuration.

%\subsection{Onboarding Implementation with TIAGo Base.}
\subsection{Robotic Platform}
For our initial tests, we integrated CARIS  with a TIAGo Base robot \cite{noauthor_tiago_nodate}. We used the onboard 2D LIDAR and the built-in movement and TTS capabilities to handle localization and speech. For video input, we attached an Oak-D camera running on a Raspberry Pi 4, which streamed footage to a FastAPI \cite{noauthor_fastapi_nodate} backend on an Apple Silicon-based laptop. As mentioned before, our computer vision algorithms were implemented using the YOLO framework, and leveraging the Metal Performance Shaders (MPS) backend available on Apple Silicon for efficient inference. The system is also compatible with CUDA-based GPU acceleration and CPU-only execution, allowing for flexible deployment across various hardware platforms.
Although we used a TIAGo Base, CARIS can be adapted to most ground-based mobile robots with basic localization and speech.

%This backend managed both computer vision tasks and LLM requests, leveraging either Gemini Flash 1.5 \cite{google_gemini_2025} API in the cloud or local models (Llama 3.1 8B \cite{meta_llama_2024}, LLaVA 7B \cite{liu_visual_2023}). This setup illustrates how CARIS modules can be split or combined as needed: while movement and TTS relied on ROS and rosbridge, the camera operated separately from the robot using WebRTC for faster streaming. Likewise, one could run TTS independently of the robot, bypassing ROS and using a simple API instead. 
%, regardless of protocol. %For example, a robot setup could run TTS on a separate device using a non-ROS API, while maintaining ROS for navigation. 
%By supporting such interchangeable modules with minimal hardware changes, CARIS accommodates a wide range of HRI scenarios for social contexts in real-world settings, though specialized platforms (e.g., drones, humanoid robots, quadruped robots, etc.) lie beyond our current scope.

\section{Pilot Study Evaluation}

CARIS is a complex WoZ system that enables a wizard to control a robot using several functionalities (see Section \ref{sec:caris}); thus, it is important to evaluate the ability of CARIS to facilitate interactions between humans and robots in different settings, usable wizard interactions using CARIS, and the effectiveness of CARIS is HRI scenarios.
We conducted an IRB-approved pilot study (\#XX) to evaluate and refine our initial system based on user and wizard feedback.

\textbf{Study Tasks:} We developed two scenarios to evaluate CARIS’s utility in various HRI contexts with one participant acted as the wizard (controlling the robot) and the other as the end-user (interacting directly with the robot).
%One scenario involved using CARIS for a \textit{tour guide robot} where the wizard introduced a robotics research lab space on a university campus, while the end-user acted as a visitor. The second scenario engaged end-users in \textit{mental health check-ups} by conducting a mental health assessment during a stationary interaction. Each scenario required collaboration with one participant acting as the wizard (controlling the robot via the interface) and the other as the end-user (interacting directly with the robot). 

\begin{itemize}
\item \textit{Room tour:} We asked the \textit{wizard} to guide the user to at least two locations using WoZ control of the robot, engaging with the multi-modal LLM capabilities, using text-to-speech for communication, and utilizing the people tracker function. The \textit{user} followed the robot on the tour and asked questions about the environment. 

\item \textit{Mental health check:} The \textit{wizard} teleoperated the robot to the user and conducted a routine mental health assessment through conversation, such as \textit{“On a scale of 1 to 10, how would you rate your stress levels today?”} and \textit{“I’ll make sure to pass along this information to the wellness team. If you need to speak with someone, they can follow up with you.”} Participants were encouraged to utilize the LLM capability and the people tracker.
\end{itemize}

\textbf{Study Design:} We used a between-subjects design, where participants for the two scenarios were different. However, one of the seven participants experienced both the wizard and end-user roles within the \textit{tour guide robot} scenario, introducing a minor within-subjects element.

\textbf{Participants:} We aimed to collect eight data points (two users and two wizards for each scenario). We used convenience sampling to recruit 7 participants near the experiment location including two females and two males, ages aged between 21 and 27 years ($M=22.75$, $SD=2.87$) Four out of seven participants completed the demographic survey. 
Participants familiarity with robots ranged from 1 to 4 ($M=2.75$, $SD=1.5$) on a 5-point scale from 1 (Very unfamiliar) to 5 (Very familiar). 

\textbf{Materials:} We deployed CARIS on two laptops: one controlling the robot and the other capturing surroundings.

%We deployed CARIS on a laptop for the wizard to control the robot, while another laptop mounted on the TIAGo robot ran the camera to capture the surroundings and interactions with the user.

%{\color{black}While the user's awareness of the wizard's presence may influence their experience, participants were made aware during recruitment that the study involved a wizard, so they already knew this beforehand. We envision exploring the differences in perceptions between wizard-controlled and fully autonomous systems as part of future work. }

\begin{figure}[t]
  \centering
  \includegraphics[width=1.0\linewidth]{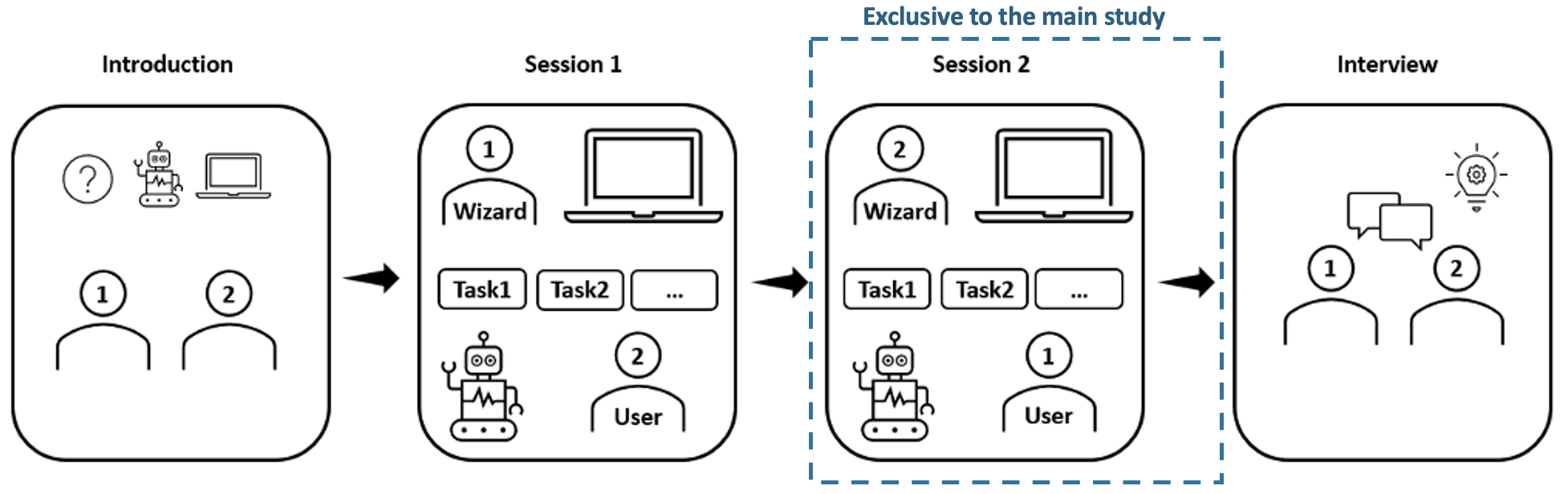}
  \caption{Procedural flow of study scenarios in our pilot and main study where participants switched roles in the main study.} %The role switch between the two participants in a given scenario is exclusive to the main study.}
  \label{fig:study-flow}
  \vspace{-4pt}
\end{figure}

\textbf{Procedure:} Figure \ref{fig:study-flow} illustrates the procedural flow in our pilot study. 
Our goal is not to fine-tune our tool to: 1) allow participants to interact with the CARIS interface to gather their perceptions of usability, and 2) collect feedback on the nuances of adaptability to their preferences and different contexts, as well as the 3) scalability of applying this fundamental tool to other use cases.
Each participant could choose from the available roles and scenarios, collaborating either with another participant or with the researcher conducting the experiment. 
 Although HRI studies traditionally involve the participants being unaware of the wizard and their role, our study focuses on the usability of CARIS and it's suitability across contexts. Futhermore, participants switched roles in our study so they inheritantly were informed about both roles.
 In the \textit{mental health check-up} scenario, the end-user and the wizard were in the same large room but positioned far apart. In the \textit{tour guide robot} scenario, the wizard guided the end-user through various spaces, and they were in the same room only when the end-user was led there. 

%{\color{black} We first introduced the participants to the study, explaining the tasks and their role in the interaction. Then, we provided the wizards with a tutorial on operating the CARIS system and gave them several minutes to practice before starting the study.}

At the beginning of each study session, we introduced the participants to the study and the robot, explaining the tasks and their role in the interaction. Then, we provided the wizards with a tutorial on operating the CARIS system and gave them several minutes to practice before starting the study. %we provided an introduction to the tasks, the robot, and the available capabilities of our interface. 
To balance participants’ subjective willingness to explore with the study’s objectives, we employed a task-oriented design. This approach allowed participants to complete tasks that encouraged interaction with the components of CARIS, rather than forcing them to follow a strict script. %ly defined procedure tailored to the scenario. 
Study participants were required to complete one session in their chosen role (i.e., user or wizard). They were welcome to try the other role in an additional session, but this was not mandatory. Among the seven participants recruited, one completed the lab tour scenario in both roles. The remaining six participants completed a session in a single role.

\textbf{Data collection and Analysis:} We measured the usability of CARIS by administering the System Usability Scale (SUS) \cite{brooke_sus_1995} to the wizards. We also conducted four semi-structured group interviews with the user and wizard to gain deeper insights into nuanced perspectives of the interface layout and capabilities. We asked participants about what they liked and disliked about CARIS and the robot. We addressed participants’ questions about broader applications of CARIS capabilities and shared the rationale behind our design, which served as a basis for brainstorming possibilities for extending the tool to different contexts and customizing it to various user preferences. We recorded audio data during interviews and we took notes when necessary. The audio recordings were transcribed and analyzed using thematic coding \cite{clarke2017thematic} to identify key themes discussed by participants.

\section{Results: Pilot Study}
\label{sec: pilot}

\textbf{SUS Results:} Four wizard participants reported the following SUS scores: 82.5, 60, 5, 12.5 ($M = 40$, $SD = 37.36$). The average positive SUS scores range from 2.25 to 3.5. The lowest positive score was for Q3: ``I thought the system was easy to use." and the highest positive score was for Q5: ``I found the various functions in this system were well integrated." Overall, the positive scores reflect a generally acceptable perception of the system's functionality and integration, but still room to improve the ease of use. 

The average negative SUS range from 3.25 to 4.25. The highest score was for Q2: ``I found the system unnecessarily complex" and the lowest was for Q10: ``I needed to learn a lot of things before I could get going with this system." 
Overall, it suggests that reducing complexity and simplifying the learning process would be a priority to improve the overall user experience. Due to the highly varied scores, we refer to the following interview results.

\textbf{Improve the robot’s movement control and dialogue capabilities:} Participants reported that controlling the robot by clicking the arrow buttons on the interface was difficult to use and often resulted in motion that is not smooth: \textit{“Sometimes you expect it to go, but it doesn’t”} P2-Wizard, which caused frustration. 
As suggested by two wizard participants, we decided to use a laptop keyboard instead of clicking arrows on the interface for more intuitive robot control. %P[tour-wizard-2], which could be frustrating. 
%As suggested by P[tour-wizard-2] and P[mental-user-2], we decided to use a laptop keyboard instead of clicking arrows on the interface for more intuitive robot control.

\textbf{Clarify and simplify Interface components layout design:} Participants reported that the CARIS interface was too crowded, making it difficult to determine the functions of the buttons. Furthermore, the cluttered layout reduced the space available for components they found more important. 
For example, P1-Wizard suggested reducing the space allocated for selecting the LLM model, and using it for the video camera capturing the user’s side: \textit{“Image model and text model [selection] doesn't have to take up as much space, and it could be a drop-down to save space because I think what people really want to see mainly is this part and the video camera.”} 
Additionally, the label for the prompt text box was noted as unclear. P1-Wizard suggested: \textit{“I like [the design that I can have notes on the side...]. Maybe changing it to ‘notes’ or a different word? Just because I think ‘extra information’ can be confusing.”} 
Thus, we replaced the LLM model selection with a dropdown menu to save space, increased the space allocated for LLM prompt/response text. We also clarified the prompt inputs (e.g. image and text), and renamed “extra information” to “notes” for clarity and intuitive understanding of interface functionalities.

\section{Main Study Evaluation}

Using lessons learned from pilot studies, we conducted a \textit{main study} to assess the usability of CARIS improvements.

\textbf{Participants:} We recruited four participants (2 females and 2 males), aged between 21 and 28 years ($M = 24.5$, $SD = 2.89$) with a robot familiarity rating that ranged from 1 to 4 ($M = 2.5$, $SD = 1.29$). 

\textbf{Study Task, Materials, and Procedure:} %Participants in the main study were different from those in the pilot study.
We used a within-subjects design, where participants acted as the wizard and user. % the two scenarios were different, indicating a between-subjects element.  
Unlike the pilot study, two participants within the same scenario switched roles after one session, requiring them to complete two sessions as both the wizard and the user, introducing a within-subjects study design (see Figure \ref{fig:study-flow}). We expected that experiencing both roles would help participants identify usability challenges and suggest improvements for smoother task flow. The study task, materials, and scenarios are identical to the pilot study.

%Other than this, participants engaged in the same study task and used the same materials for both scenarios as the pilot study.% illustrates the procedural flow of each scenario in our main study.

\textbf{Data Collection and Analysis:} Data collection is consistent with the pilot study. The interviews were conducted with wizard and end-user participants simultaneously to encourage collaborative discussion. We conducted two interviews, each with two participants. The interviews were recorded, and we took notes when necessary. The recordings were later transcribed and analyzed using thematic coding \cite{clarke2017thematic}.

\section{Results: Main Study}

\textbf{SUS Results:} The SUS scores reported by the four wizard participants ranged from 22-65 ($M = 43.75$, $SD = 19.63$). The positive SUS questions resulted in average scores ranging from 2.5 to 3. ``I would imagine that most people would learn to use this system very quickly." (Q7) resulted in the lowest average score and ``I thought the system was easy to use." (Q3) resulted in the highest average score. %The positive SUS questions resulted in average scores that ranged from 2.5 and 3. The lowest average score was for ``I would imagine that most people would learn to use this system very quickly." (Q7), and the highest was for ``I thought the system was easy to use." (Q3) and “I felt very confident using the system.” (Q9). The standard deviations are between 0.96 and 1.83, with Q9 being the highest, and ``I think that I would like to use this system frequently." (Q1) and  “I found the various functions in this system were well integrated” (Q5) being the lowest.  
Overall, the positive scores suggest that participants generally found the system easy to use and felt confident using it, but there is room for improvement in terms of ease of learning and function integration. 
The negative SUS questions resulted in averages ranged from 2.5 to 3.75. The highest average score was for Q6: “I thought there was too much inconsistency in this system.” and Q8: ``I found the system very cumbersome to use." %The negative SUS questions resulted in averages ranging from 2.5 to 3.75. The highest average score was “I thought there was too much inconsistency in this system.” (Q6) and ”I found the system very cumbersome to use." (Q8). 
In contrast, the lowest average score was 2.5 for ``I’II needed to learn a lot of things before I could get going with this system." (Q10). %The standard deviations are between 0.5 to 1.73, with ``I think that I would need assistance to use this system." (Q4) being the highest, and Q6 and Q8 being the lowest. Overall, the negative scores indicate that while participants did not feel the system was overly difficult to learn, some inconsistencies and usability issues may have affected the overall experience. %In general, responses from both the pilot and main studies are mixed. %, which may partly reflect differences in participant evaluation criteria, suggesting the need for more detailed follow-up interviews to clarify these variations.
These results demonstrate moderate usability CARIS.
However, the self-report results do not fully capture participants' experiences. To understand these nuances, we further analyzed the interviews using thematic coding \cite{clarke2017thematic}. 
We identified areas of improvement, gained insights into how the capabilities are perceived across different contexts, and developed a better understanding of what adaptability means in a robot control interface. %We identified areas for improving usability in the interface layout, gained insights into how the capabilities are perceived across different contexts, and developed a better understanding of what adaptability means in a robot control interface.

%\textit{Smoother Integration between Movement and Communication:} Overall, participants found it easy to move the robot using the keyboard and navigate it using the video stream. As P2 noted, \textit{“I thought the overall navigation, voice, and map were very well done and made navigation easy.”} Similarly, P1 mentioned, \textit{“[As for] controlling the robot and seeing where it was going, the camera was really good. I got a good field of view on where I should click to navigate, so I didn’t hit anything.”} However, while the LIDAR view provided additional information for some participants, it confused others. For instance, P3 found the LIDAR map somewhat confusing to interpret initially and suggested that a \textit{“water drop”} symbol, commonly used to indicate positions (e.g., similar to Google Maps), could make it more interpretable. 

\textbf{Smoother Integration between Movement and Communication:} Wizard participants suggested reducing interruptions between moving the robot and communicating with users. 
Wizards found that moving the robot while typing to communicate with the user difficult. 
As P4 noted, \textit{“I feel like the arrow keys are linked to the keyboard [for moving] and the keyboard is also linked to this [typing] task.”} P2 added, \textit{“Allowing for multiple actions to happen at once would make this much more user-friendly. [Wizards should] be able to type, send, and move simultaneously.”} %To address this issue, participants suggested using different key configurations or separate devices for movement. For example, P3 suggested, \textit{“If it can be separated into other forms, these keys won't be an interruption.} %[...] Or you can just set [the keys for moving] to ‘WASD’.”} 
%Another suggestion was, \textit{“Moving the robot around is quite intuitive with the keyboard arrows but would probably work better with a joystick.”}

%\subsection{A Clearer Separation between the LLM Capabilities and the Communication Module Functionalities}
%\textbf{A Clearer Separation between the LLM Capabilities and the Communication Module Functionalities:} Participants suggested a clearer separation between the LLM module and the communication module in terms of both layout and functionality. For example, we implemented functions like \textit{“saying hello”} and \textit{“waiting for LLM response”} when the LLM is generating responses, helping the user understand that the LLM is processing rather than not responding, but participants perceived these as more aligned with communication rather than LLM capabilities. P1 agreed with P2, stating, \textit{“In my opinion, it should be something more like: down here is all the communication, and up here is all the LLM stuff.”} They further noted, \textit{“This (greeting) requires no LLM usage, while this (LLM prompts) does, but they seem to be confusingly mixed.”}

\textbf{A Clearer Separation between the LLM Capabilities and the Communication Module Functionalities:} The wizards had different communication preferences, leading to suggestions for varying prioritization of the two modules. For instance, P2 considered typing to be the most important mode of communication during the room tour session as a wizard and, therefore, suggested prioritizing the text frame used for typing. P2 explained, \textit{“I found the most use in just typing, so I believe prioritizing this is important.”} P2 added, \textit{“It depends on the task because, for most situations, if I’m communicating with someone, having an LLM say what's there is not helpful. If I'm trying to communicate, I would rather just type it myself. Typing should be the first priority.”}

\textbf{Recommended LLM Prompts and Commonly Used Phrases Based on the Context:} Participants suggested that CARIS automatically generate suggested prompts based on what the user is saying. This feature would help reduce the wizards' workload and response time by eliminating the need to manually write LLM prompts as mentioned by P3, \textit{“There may be a few [prompt] suggestions specifically to respond to the context that was just recorded. You don't need exact words, but I do need some suggestions on how to respond.”} 

\textbf{Adaptability Across Contexts:} Participants emphasized the interoperability of CARIS to adapt based on specific contexts and tasks, with considerations for functions, layout, access, restrictions, and priorities. %In addition to the previously mentioned prompt suggestions and frequently used texts, participants noted that the prioritization and necessity of capabilities vary across different contexts. 
For instance, a mental health check scenario requires a higher level of consideration such as privacy and data sensitivity: \textit{“I may not use the take picture function because I don’t need to, or maybe I am not allowed to” (P3)}. However, in lab tour scenarios, the picture function could be essential for helping the language model provide more accurate answers about the environment. Moreover, in response to the questionnaire item, “How intuitive was it to use the language model for generating responses?” Lab tour scenarios received an average score of 3.5 on a 1–5 scale, whereas mental health check scenarios averaged only 2. This suggests a more cautious approach when using LLMs in healthcare settings compared to more straightforward introduction tasks e.g., lab tour.

%\subsection{Potential and Application}
\textbf{Future Potential of CARIS for HRI:} Participants recognized the robot's broad potential, noting that \textit{“it just needs a smooth path to go along” (P4)}. P3 envisioned this remote control role as particularly effective \textit{“when they’re in a secured building without any bumpy roads”}. They also highlighted that an adjustable camera height could broaden its applications, with P4 stating, \textit{“Like going up or down, any sort of height control. It’s like an elevator.”} P2 added, \textit{“Raising the laptop higher may improve functionality by lifting it off the floor”}. From the users’ perspective, participants suggested that \textit{“making it obvious that the robot is talking to you would be nice (e.g., eyes)” (P3)} and noted that \textit{“having the laptop camera and screen was a great way to get an idea of what the robot (and also the wizard) was seeing” (P2)}.

\section{Discussion and Future Work}

We introduced, CARIS, a modular context-adaptable WoZ system, and demonstrated its potential to evaluate HRI strategies in tour guide and mental health check scenarios. 
Future directions include 1) extending the SLAM pipeline to autonomous or semi-autonomous navigation, 2) refining navigation and conversation modules for smoother multitasking, 3) linking LLM outputs directly to robot actions (e.g., “go to the kitchen”) to reduce the wizard’s workload, and 4) using data recorded to develop data-driven models for HRI. 
We hope that CARIS will increase HRI research productivity, through new datasets, data-driven techniques, and rapidly develop socially-aware robots.

%CARIS proved effective for adaptable WoZ control but highlighted key usability issues, including cumbersome navigation-communication multitasking and a crowded interface. Future directions include refining navigation and conversation modules for smoother multitasking (e.g., joystick or separate key bindings), adding toggles to disable unneeded features, and improving icons for clearer scenario-based layouts. Another possible advancement is linking LLM outputs directly to robot actions (e.g., “go to the kitchen”), as well as on-board STT for end-users to reduce the wizard’s workload. Extending the SLAM pipeline for real-time mapping and testing CARIS with different robot onboardings or simulators can further validate it as a robust, context-aware platform bridging low-level control and advanced LLM-driven social interactions.

\bibliographystyle{IEEEtran}
%\bibliography{references_zotero}
\bibliography{references.bib}
\end{document}